\documentclass[11pt,letterpaper]{llncs}
\usepackage[utf8]{inputenc}
\usepackage{amsmath}
\usepackage{amssymb}
\usepackage{graphicx}
\usepackage{colortbl}
\usepackage[table,xcdraw]{xcolor}
\usepackage{tabularx, booktabs}
\usepackage{multirow}
\usepackage{marvosym}
\usepackage{url}
\usepackage{fancyhdr}
\usepackage{hhline}
\usepackage{authblk}

\usepackage{hyperref}

\makeatletter
\newcommand{\printfnsymbol}[1]{%
  \textsuperscript{\@fnsymbol{#1}}%
}
\makeatother

\usepackage{geometry}
\newcommand{\keywords}[1]{\par\addvspace\baselineskip
\noindent\keywordname\enspace\ignorespaces #1}

\fancypagestyle{firstpage}{\fancyhf{}
\setcounter{page}{1}
\rfoot{\thepage}
}

\begin{document}


\title{\LARGE{Deep-learning coupled with novel classification method to classify  the urban environment of the  developing world}}


%
%
%

\author{\normalsize{\thanks{Equal contribution}Qianwei Cheng\inst{1} \and   \printfnsymbol{1}AKM Mahbubur Rahman\inst{2,3}
Anis Sarker\inst{2} \and Abu Bakar Siddik Nayem\inst{2} \and Ovi Paul\inst{2} \and Amin Ahsan Ali\inst{2,3} \and M Ashraful Amin\inst{2,3} \and Ryosuke Shibasaki\inst{1} \and  \printfnsymbol{1}Moinul Zaber\inst{3,4,5}} }

\institute{\normalsize{The University of Tokyo, Tokyo, Japan \and The Independent University Bangladesh, Dhaka, Bangladesh \and The Data and Design Lab, Dhaka, Bangladesh \and University of Dhaka, Dhaka Bangladesh \and United Nations University, E-government Operating Unit (UNU-EGOV), Guimarães, Portugal}}

\maketitle

\begin{abstract}
Rapid globalization and the interdependence of the countries have engendered tremendous in-flow of human migration towards the urban spaces. 
With the advent of high definition satellite images, high-resolution data,  computational methods such as deep neural network analysis, and hardware capable of high-speed analysis; urban planning is seeing a paradigm shift. Legacy data on urban environments are now being complemented with high-volume, high-frequency data. However, the first step of understanding the urban area lies in the useful classification of the urban environment that is usable for data collection, analysis, and visualization. In this paper, we propose a novel classification method that is readily usable for machine analysis and it shows the applicability of the methodology in a developing world setting. However, the state-of-the-art is mostly dominated by the classification of building structures, building types, etc., and largely represents the developed world. Hence, these methods and models are not sufficient for developing countries such as Bangladesh where the surrounding environment is crucial for the classification. Moreover, the traditional classifications propose small-scale classifications, which give limited information, have poor scalability and are slow to compute in real-time.  
We categorize the urban area in terms of informal and formal spaces and take the surrounding environment into account. $50\, km\, {\times}\, 50\, km$ Google Earth image of  Dhaka, Bangladesh was visually annotated and categorized by an expert and consequently, a map was drawn. The classification is based broadly on two dimensions the state of urbanization and the architectural form of the urban environment. Consequently, the urban space is divided into four classifications: 1) highly informal area 2) moderately informal area 3) moderately formal area and 4) highly formal area. For semantic segmentation and automatic classification, Google's DeeplabV3+  model was used. The model uses the Atrous convolution operation to analyze different layers of texture and shape. This allows us to enlarge the field of view of the filters to incorporate a larger context. Image encompassing $70\%$ of the urban space was used to train the model and the remaining $30\%$ was used for testing and validation. The model can segment with $75\%$ accuracy and $60\%$ Mean Intersection over Union (mIoU).

\end{abstract}


\section{Introduction}

Rapid globalization has engendered a large number of people in-flow into cities.  More than $50\%$ of the global population lives in urban areas, especially in developing countries. Therefore, the sustainable development goals of the United Nations emphasize the demand for information about the physical environment of urban cities \cite{UnitedNations2015}.  Categorizing the urban area is a fundamental component of any decision making process for any kind of urban development. Classification of the urban building structures offers several benefits for the policymakers. For example, by pointing out the educational institutes, we can understand people’s access to education in that specific area \cite{Tingzon_I_2019}.  With building classifications mapped to the geo-spatial area, new educational institutions might be proposed in the desired area. Additionally, distinguishing industrial area from residential areas can help the planners to build infrastructures to manage industry wastes, water pollution, and air pollution \cite{Roy1997}.

The spatial structure and pattern of the urban environment are diverse and complex\cite{Tauben2016}. However, most of the early studies on urban classification were based on a single dimension: structural characteristics: the types and conditions of individual buildings, and building density, building height, building scale, or physical site characteristics. Many pieces of research are devoted to the analysis of cities on a spatial scale. For example, Herold et al.\cite{Herold2005} combine satellite images with urban spatial scales to simulate urban growth and land-use change. Huang et al.\cite{Huang2007} calculated 7 spatial metrics by using the satellite images of 77 cities and obtained 5 different properties of urban morphology: complexity, centrality, compactness, porosity, and density respectively. 



A good number of studies have used deep learning to categorize urban areas from satellite images. Oshri et.al.\cite{DBLP:journals/corr/abs-1806-00894} used Afrobarometer \cite{afrobarometer} survey data along with satellite images from Sentinel 1 and Landsat 8 to assess the environmental quality in Africa. They have also used OpenStreetMap. They have used deep learning technology to identify the presence of different infrastructures such as electricity, sewerage, piped water, and road. They have categorized the habitat environment by identifying the above-mentioned structures.


Neal et al. \cite{DBLP:journals/corr/XieJBLE15} have used Convolutional Neural Networks (CNNs) to analyze satellite images for luminosity at night (“nightlights”) to estimate economic activity in different areas. Then they have predicted the poverty levels in the targeted area from the estimated economic activities. 
Piaggesi et al. used \cite{Piaggesi_2019_CVPR_Workshops} a pre-trained fully convolutional variant of VGG, fine-tuned with nightlights intensity levels, baseline VGG and Resnet-50 initialized with ImageNet were used to predict income of each household. However, very few researches have performed automatic building classification based on the buildings and their surrounding environments. Manual classification of a small area has been performed in urban research. But it is not scalable for a large metropolitan city like Dhaka. Therefore, we propose a novel classification and deep learning method for automatic semantic segmentation of the geospatial images of the city.


In the early study of the classification model, researchers considered the context of developed countries that is not suitable for developing countries. Moreover, in the study of those urban classifications, the surrounding environment and building structure are not considered together. But in many developing countries with complex urban issues, apart from complex land ownership, urbanization and buildings are also multifaceted. The building structure is often closely related to the surrounding environment.
\\
In this study, we have designed a method to classify the urban environments. This classiﬁcation has the following advantages:
\begin{enumerate}
    \item It sets up the urbanization process and architectural form as two different axes instead of a single  index. Then the physical environment of the city is divided into 16 classifications. Through this classification method, all kinds of urban environmental forms can be summarized.

    \item Through manual operation, a large number of high-precision urban classification data are obtained that makes the deep learning possible.
    \item The scale of this study is large (more than $20\, km\, {\times}\, 20\, km$, less than $50\, km\, {\times}\, 50\, km$). Previous studies were limited to small-scale classification, poor scalability, and inability to provide a large amount of information.
\end{enumerate}

Because of its large scale, this classification method can be used to compare the level of cities with the data of topography, area, population density, income, education, etc. as the basis of urban geography and urban poverty research. One can complete a morphological catalog or even a list of global cities.
The method of visual recognition is used to detect and classify the physical environment of many cities and draw the map, and then the deep learning algorithm is used to train the map data set. Then, we use the trained deep learning algorithm to identify the building classifications in the unseen test data. Particularly we have developed a transfer learning method to semantically segment the satellite image of Dhaka city into four different urban areas automatically. The areas are segmented based on the buildings and their environment. Specifically, we have used a transfer learning method to categorize the area into 1) highly informal area; 2) moderately informal area; 3) moderately formal area and 4) highly formal area. Unlike other urban classification models, we propose the novel urban classification technique that includes the buildings as well as their surrounding environment. Actually, the texture and shape differences among various building structures are present in the satellite images. These textures and shapes can be learned by the state of the art deep learning models.  Consequently, the deep learning models would be able to capture the texture and shape differences and can segment the satellite images according to the proposed classifications. In this paper, we have used google’s Deeplabv3+ model. The model has been trained with $70\%$ area of the Dhaka city and tested on the rest $30\%$ area. The DeepLabv3+ model has used convolutional operations to analyze the textures and shapes in various scales. Therefore our contribution includes 1)  Novel urban classification, 2) Segmentation with Deep Learning.

In the next section, we discuss the theoretical background of the urban building classification and deep learning-based automatic methods. Then we describe our novel classification model with sufficient examples. In the fourth section, we discuss the deep learning method to semantically categorize the building classifications automatically from satellite images. Then in the fifth section, we have presented our results followed by the conclusion.

\section{Research Background}
\subsection{Urban Area}

Global economic, social, and political circumstances have transformed  the urban and rural in such a way that it has become very difficult to define urban and rural area in a distinguishable way. However, here, we intend to define the urban-rural area in the context of Bangladesh.
An urban area can be defined based on population density, the concentration of administrative bodies, infrastructure, and a diverse set of livelihood and income generation activities. Additionally, urban areas must be characterized by high population density when compared to other areas.  In an urban setting, the forms of livelihood and income generation activities will be diverse and unlike rural areas not bound mainly to agricultural production \cite{Rana2011}. Some cities might be defined by municipal boundaries. But many of the urban areas are usually characterized by the presence of administrative structures such as government offices and courts and a relative concentration of services such as hospitals and financial institutions such as banks.
\subsubsection{Urban area of the developing world}
Urbanization, generally, refers to an increasing shift from agrarian to industrial services and distributive occupations. $50\%$ of the world’s population lives in urban areas, and this proportion is expected to increase to $68\%$ by 2050. By 2050, the increase in world population may add another 2.5 billion people to urban areas [59]. 
For example, between 2003 and 2010, the city of Hyderabad in India experienced a $70\%$ increase in slum areas \cite{kit2013}. Unfortunately, the situation is not unique to India or the global south. UN estimations show that the global slum population is around one billion at present and may rise to two billion by 2030 (UN Habitat Report 2016\cite{UN-Habitat2016}). 
In developing countries therefore understanding the socio-economic condition of different regions of a city can be assisted by the identification of different formal and informal settlements. Therefore, there has been a lot of focus on the identification and mapping of urban areas of a city. 
\subsubsection{Urban area in Bangladesh}
In 2019, the urban population for Bangladesh was $37.4\%$. The urban population of Bangladesh increased from $7.6\%$ in 1970 to $37.4\%$ in 2019 growing at an average annual rate of $3.33\%$ \cite{Siddiqy_2017}. 
The trend of present urbanization poses a huge challenge for the sustainable metropolitan city in Bangladesh \cite{Rana2011}. 
The infrastructural services, basic amenities, and environmental goods, environmental degradation, extensive traffic and accidents, violence, and socioeconomic insecurity are the major challenges for the city. Generally, these challenges are very common in highly informal and moderately informal areas. Distinguishing these areas from the formal area on a large scale is the major challenge for a mega city like Dhaka. 

\subsubsection{Urban environment classification}

\begin{figure}
\centering
\includegraphics[scale=0.6]{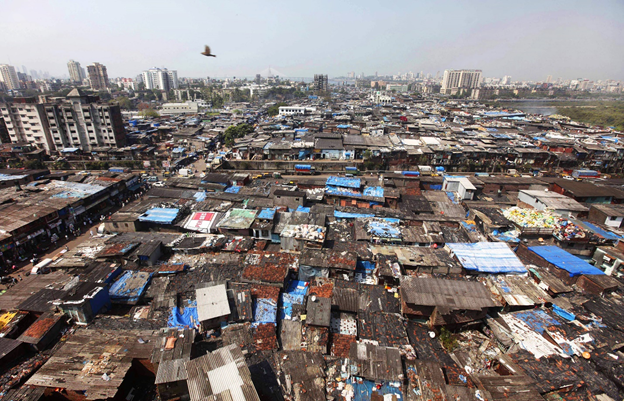}
\caption{The morphological appearance of the urban environment}
\label{fig1}
\end{figure}
In the urban environment, different regions have different classifications (Figure \ref{fig1}). They are composed of different physical appearances, such as green spaces, artificially constructed and naturally formed roads; regular, random, complex and dense buildings, etc.
The study of urban morphology classification can provide important reference for the  evolution and boundary division of historical regions, and provide in-depth understanding and training for the practical application fields closely related to urban development, planning, architecture, urban management, public policy, etc. Barke et al. \cite{Barke2018} approach understanding urban form and its complexity, linking urban form parameters with its spatial performance, is an important way for a city to develop well. Kärrholm et al. \cite{Karrholm2008} and many other authors point out the importance of understanding urban morphology as a key element of urban sustainable development. 
Kevin Lynch's redefined urban environment includes a spatial relationship of urban physical characteristics (natural physical form, established physical form) \cite{Lynch1984}. Since the 1990s, with the development of geographic information system, more and more people understand the complexity of the urban environment from a comprehensive perspective.

In the urban environment, a consensus is there on the core basic elements of the form of solid buildings: streets, plots, buildings, etc. Urban form can combine design, shape, and clarity from the courtyard to cluster scale. In the methods of urban morphology topology and historical geography, the concept of composition level, which focuses on specific aspects of architectural form, has been fully developed  \cite{samuels2012urban}. Jwr whitehand extends Conzen's theory and establishes a historical geographic approach \cite{Oliveira2018}.
Urban classification is the classification of three core elements of streets, plots, and buildings in a complex urban environment \cite{Tauben2018}. In general, the type and conditions of individual buildings \cite{Jain2007} and the spatial layout of these buildings \cite{gueguen2014classifying} define the urban environment. ISA baud et al. \cite{Baud2010} used physical indicators: layout structure, building density, building scale, and other physical site characteristics, the building types are divided into four classifications. According to Taubenbóck et al. \cite{Tauben2018}, firstly, cities are classified according to the physical density of buildings. Then, in the study of the classification and characteristics of settlement form, the characteristics of settlement form of five cities are extracted from the three core elements of building density, building orientation, heterogeneity index, building scale, and height. Finally, the five cities are divided into nine classifications. Slums, similar to slums, mixed unstructured and unstructured communities, mixed structured and unstructured communities, and structured and formally planned communities \cite{Tauben2016}.


\subsection{Machine Learning for understanding Satellite image}
There are hundreds of Earth observation satellites providing data at different spectral, spatial, and temporal resolution. ERS 1 and 2, Sentinel 1 and 2 satellites launched by European Space Agency (ESA), Landsat 1-8 satellites maintained now by US National Oceanic and Atmospheric Administration (NOAA) have been providing satellite data for free and have initiated research in this field. Commercial satellites such as IKONOS, RapidEye, and WorldView 1-4  provide higher spatial $({\sim}5\, meter/pixel)$ resolution) satellite imagery compared to that of Landsat and Sentinel (tens to hundreds of meters). The resolution means the actual ground distance (in meter) that has been sampled by one pixel. It is also called ground sampling distance. Microsoft corporation has launched a couple of satellites (Bing Satellites) to provide remote imagery with high resolution ($2.22 meter/pixel$).  

As these high-quality images have visual channels (RGB), researchers apply spatial filters to analyze the textures and shapes of the satellite images to identify forests, water, buildings, farmlands, and meadows. In \cite{Bruzzone2006}, the authors proposed a novel feature-extraction block extract spatial context of each pixel according to a complete hierarchical multilevel representation of the scene. Then they classify the features with SVM.  Cheng et al. \cite{Cheng2015} discuss traditional features such as histogram of ordered gradients(HOG), scale-invariant  feature transform(SIFT), color histograms(CH), and local binary patterns(LBP) etc. Researchers use these features to segment the satellite images for different land covers by exploiting the support vector machine(SVM).
However, due to the high volume, heterogeneity of satellite data, nowadays, deep learning methods are getting more preference over traditional manual or statistical methods to explore these data sets \cite{DBLP:journals/corr/abs-1709-00308},\cite{Nogueira_2017}. These methods have been instrumental to explore such data within reasonable time and cost.  Deep learning methods show excellent performance in classifications, image segmentation, as well as semantic segmentation \cite{Abs2019}. The emergence of deep nets also help  segmenting high-resolution satellite images. 
\subsection{Semantic Segmentation for building classification}


\begin{figure}
\centering
\includegraphics[scale=1.0]{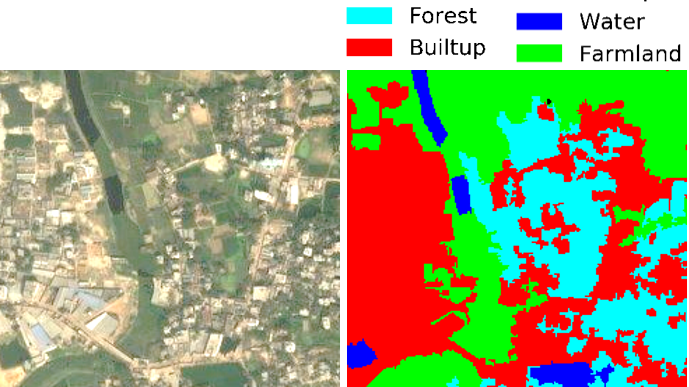}
\caption{a. Satellite image b. Land use land cover segmented image}
\label{fig2}
\end{figure}

\textbf{Semantic segmentation on satellite data:} Satellite images have multiple channels including the visual ones: red, green, and blue (RGB). The RGB image has the texture and shape information of the ground. Since texture and shapes can be captured by deep learning-based semantic segmentation, remote sensing researchers try to segment specific areas based on different classification tasks. Therefore, they have approached remote sensing tasks through the use of deep learning methods. The classification tasks might range from land use land class to slum detection, from tracking urban expansion to tracking forest degradation, etc. 

\textbf{Semantic segmentation  for urban environment development:} More recently, machine learning-based automated analysis of satellite imagery has been used to understand human socio-economic conditions at a local (city, village, or country level) as well as global level. Both deep learning methods (e.g., Convolutional Neural Networks (CNN)) and shallow learning methods (e.g., Support vector machines and Random Forests) are being extensively used in predicting socio-economic indicators using daytime optical satellite imagery, radar data, and Night Time Light (NTL) emission captured by satellite sensors. However, to train such models one needs both large quantities of satellite images, as well as non-satellite data from different sources such as census, household and agricultural surveys, administrative data, and other environmental and meteorological data \cite{Andries2018}. Machine learning methods have been used to estimate crop yields \cite{10.5555/3298023.3298229}, land cover and land cover changes \cite{Nogueira_2017}, slum detection \cite{Mahabir2018}, mapping urban growth predicting and mapping poverty \cite{Benza2016}, population estimation \cite{DBLP:journals/corr/abs-1708-09086}, estimation of GDP etc. 

Authors in \cite{Andries2018} provides a comprehensive review of the potential use of earth observation data (both satellite and aerial imagery) to estimate socio-economic indicators set by the UN's Sustainable Development Goals and the `proxy' indicators. In the following we discuss the major directions that research has been conducted in this field of study, focusing on Bangladesh's context.
Various CNN’s have been used to classify land use and land cover in many different types of satellite image datasets[UC Merced \cite{UCmerced2010}, OSM Land Use Data, Climate Change Initiative (CCI) \cite{cci2013}, Global Land Survey etc.] in addition with other tabular and/or temporal data. Oshri et.al. \cite{DBLP:journals/corr/abs-1806-00894} used Afrobarometer \cite{afrobarometer} survey data along with satellite images from Sentinel 1 and Landsat 8 to assess infrastructure quality in Africa. They propose multi-channel inputs that utilize all available channels from respective satellites instead of using only the RGB bands. Pretrained Imagenet weights have been used for the RGB channel and the novel channels are initialized with Xavier Initialization. A Residual Network based architecture (ResNet101) has been used with some minor modifications in their work which achieves better classification performance than Nightlights \cite{Jean790} based works, OSM, and some other baseline methods. 
While predicting poverty from satellite image as a socio-economic indicator of city areas, images from DigitalGlobe\cite{digitalglobe} platform and the google maps have been used as input imagery where the spatial resolution used was $.5\, m/pixel$ and $2.5\, m/pixel$ respectively \cite{Piaggesi_2019_CVPR_Workshops}. A pretrained fully convolutional variant of VGG fine tuned with nightlights intensity labels, baseline VGG and Resnet-50 initialized with ImageNet has been used to predict income of  each household. The ground truth data has been produced from EOD (Encuesta Origen Destino de Viajes), a mobility survey realized from July 2012 to November 2013 by assignment of the Chilean Ministry of Transport and Telecommunications\cite{eod}. Resnet-50 performed better overall yielding slightly better accuracy. 

\section{Detail Description of Novel urban classification}

It is generally accepted that urban areas and buildings are the basic types for analyzing urban forms.
 Frey et al. have pointed out that the structure and form of traditional cities developed slowly and gradually without formal planning and design.
Marshall \cite{marshall2009cities}) also pointed out that the physical characteristics of streets, plots, and buildings are different. An important part of seeking to understand how cities develop and evolve is to determine the specific combination of these elements.
In ordinary human settlement planning, roads and blocks are planned first, and then buildings are constructed according to the divided blocks. On the other hand, in unplanned areas, buildings are planned first, and then the streets are planned. But the formation of towns is not limited to these two modes. There are two other more mixed modes, in which unplanned buildings are built in the planned road network, and then roads naturally appear in this unplanned building group; Construct planned buildings in areas where the road network is not planned. 

\begin{figure*}
\includegraphics{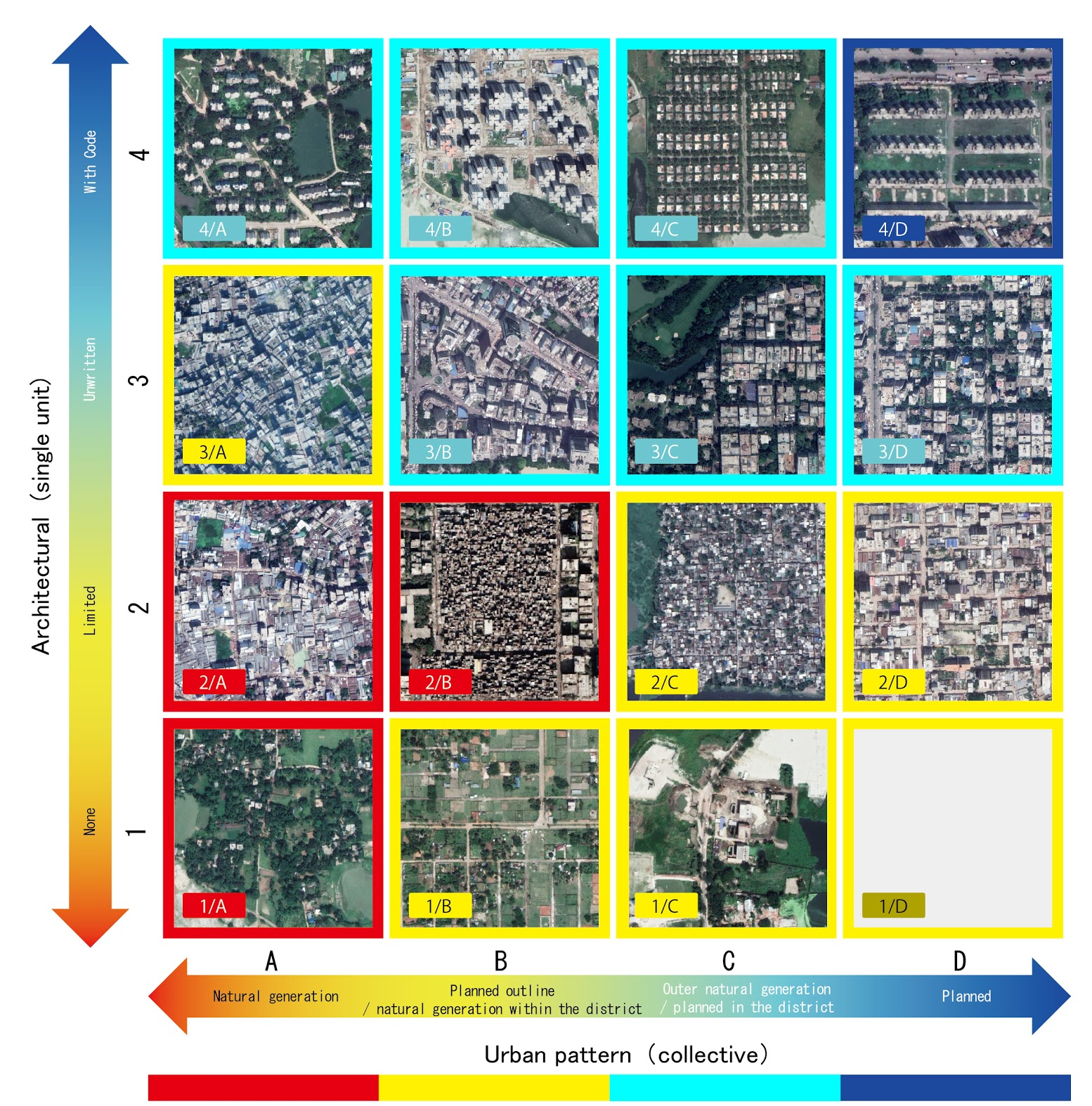}
\caption{Urban environment topology}
\label{fig3}
\end{figure*}

Based on the concept mentioned above, in this study, the physical environment of the city is classified by two indicators: diversity of buildings and street patterns (aggregate). The diversity of buildings is expressed as a single building, and it is divided into four categories. The fully planned buildings which in the same code are deﬁned as ``with code", and the number is marked (4)$;$ Untyped buildings are defined as ``unwritten", and the number is marked (3), It is usually expressed in historical buildings, old blocks, factories, public buildings, etc; The use of cheap materials, inferior buildings with technical and economic constraints, and self-construction are defined as ``limited", and the number is marked (2); High-quality housing in the suburbs and rural buildings, which have no technical, economic, or social restrictions, are defined as ``none" and marked as a number (1). 
Building types with codes and unwritten rules are relatively easy to distinguish, but the criteria for distinguishing building types with technical and economic constraints are not designed according to a prescribed layout. While its form looks random, it uses temporary, irregular, and diverse building materials. And these forms of buildings are land-risk slopes, lowlands, along rivers, around industrial areas may be identified. A type of ``none" indicates a state when a building type having technical and economic constraints is free from the limitation.
For another indicator, street mode (collection), the planned street network is defined as ``planned" and marked with a letter (D); naturally generated roads are defined as ``natural generation" and marked with a letter (A); in addition, According to the regularity of the street pattern, the outside is the planned street network, and the inside is the naturally generated road network. The type is defined as ``planned outline + local natural generation", marked with the letter (B); the external road network is natural Generated, the planned type of the internal road network is defined as ``Outer natural generation + Planned in the district", marked with a letter (C).

After that, the four types of indicators ``diversity of buildings" (marked with numbers 1, 2, 3, and 4 from the bottom of the vertical axis) and the four types of indicators ``street patterns" (marked on the left side of the horizontal axis as The combination of letters A, B, C, D), the combination of two indicators forms a total of 16 types. The obtained matrix is used as a standard criterion for categorizing every urban area with a scale of $400\, m{\times}400\, m$. According to this classification method, we classify the urban environment of Dhaka City, Bangladesh(Figure \ref{fig3}).



In cities in developing countries, it is often possible to distinguish between formal areas characterized by appropriate spatial planning and facilities (buildings, green space, road infrastructure) and informal areas without urban development. Informal areas are usually composed of irregular and restricted buildings, located near dangerous land, industry, garbage, swamps, and flood-prone areas [60].

Considering the relationship between urban environment type and formal/informal degree, three types of 2/A, 2/B and 1/A located at the lower-left pole are the street pattern of ``natural generation (A)". Alternatively, it is inferred that the street pattern of ``planned outline + natural generation in the district (B)" contains ``limited (3)" buildings, and the degree of formalness is extremely high. This classify was Color1. At the top right pole is a 4/D where the same street is lined with buildings with the same code on the planned street. The remaining $12$ types are assumed to be informal degrees, and the 1/B, 1/C, 1/D, 2/C, 2/D, and 3/A groups are indicated by Color2. The 3/B, 3/C, 3/D, 4/A, 4/B, and 4/C groups were referred to as Color 3 below and were classified into four types of urban environment types. The selection colors of urban environment topology is shown in Table \ref{table1}.

\begin{table}
\centering
\caption{Selection of colors for urban environment topology}
\label{table1}
\begin{tabular}{|c|c|l|} 
\hline
\begin{tabular}[c]{@{}c@{}}Urban Environment\\Topology \end{tabular}        & Classification    &   \multicolumn{1}{c|}{Color}  \\ 
\hline
\begin{tabular}[c]{@{}c@{}}Red\\ (1/A.2/A.2/B) \end{tabular}                & Highly Informal   & {\cellcolor{red}}   \\ 
\hline
\begin{tabular}[c]{@{}c@{}}Yellow(1/B.1/C.\\1/D.2/C.2/D.3/A) \end{tabular}  & \begin{tabular}[c]{@{}c@{}}Moderately  Informal \end{tabular} & {\cellcolor{yellow}}           \\ 
\hline
\begin{tabular}[c]{@{}c@{}}Cyan(3/B.3/C.\\3/D.4/A.4/B.4/C) \end{tabular} & \begin{tabular}[c]{@{}c@{}}Moderately Formal \end{tabular}  & {\cellcolor{cyan}}           \\ 
\hline
\begin{tabular}[c]{@{}c@{}}Blue\\ (4/D) \end{tabular}                    & Highly Formal                                                     & {\cellcolor{blue}}          \\
\hline
\end{tabular}
\end{table}

\begin{figure}[!ht]
\centering
\includegraphics[scale=0.55]{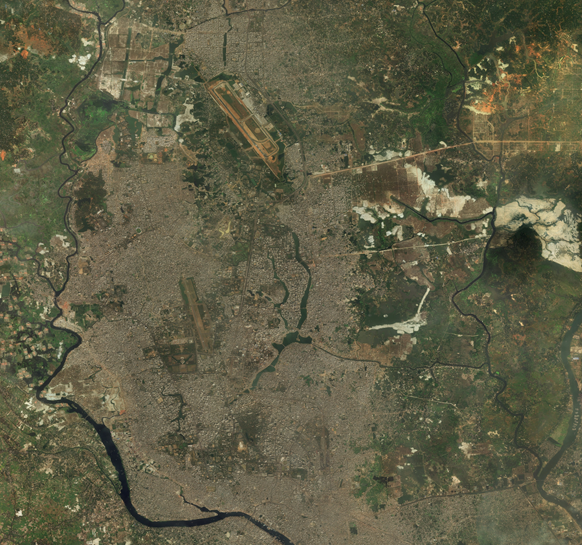}
\caption{ Input Satellite image for Dhaka city - Source: BING satellite imagery }
\label{fig4}
\end{figure}




\section{Automatic Building-classification using Deep Learning}
\subsection{Data Collection, Annotation and Preprocessing}
\begin{figure}[!ht]
\centering
\includegraphics[scale=0.25]{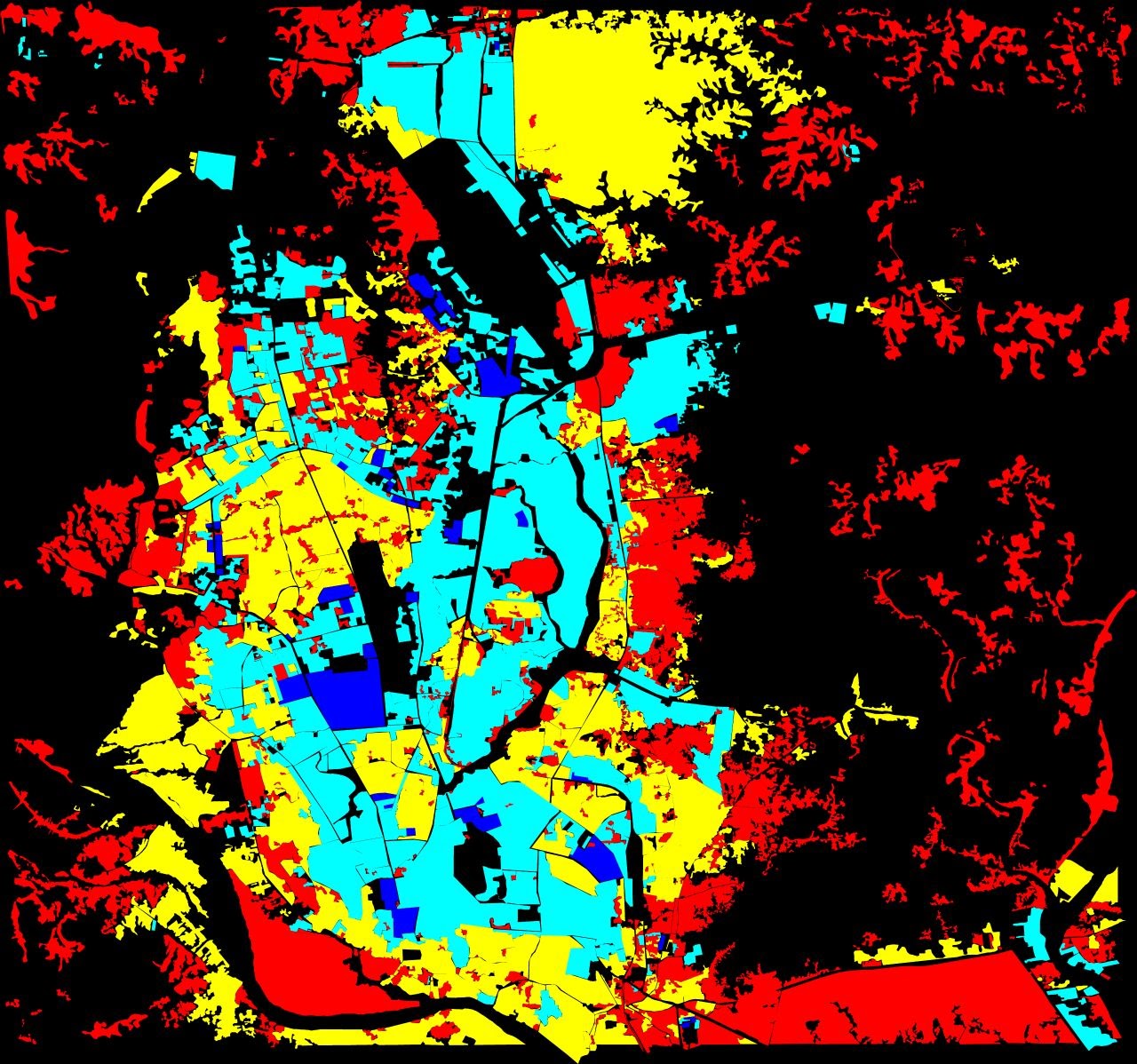}
\caption{Ground truth for building classification. Red: Highly informal area (Arrival city); Green: Moderate Informal Area; Cyan:  Moderate Formal Area; Blue: Highly Formal Area ; Black: Unrecognized Area}
\label{fig5}
\end{figure}

We use the freely available Bing satellite imagery for our building classification task. The Microsoft Corporation provides high-resolution imagery for public uses from the Bing satellites.  Satellite image of the Dhaka metropolitan area has been downloaded at a zoom level of 17 which has a spatial resolution of $2.26\, {m}/{pixel}$. Even though the city administrative boundaries are not of a perfect rectangular shape, we produce an estimated rectangular shapefile of Dhaka city by using its respective administrative shapefile for the sake of simplified grid preprocessing.  Images are downloaded with geolocation metadata so that further analysis by geo-referencing can be done later. ArcMap GIS Software has been used to superimpose the shapefile of rectangular boundary and crop the region of interest from downloaded image(Figure \ref{fig4}) and the ground truth(Figure \ref{fig5}) image respectively.

It can be seen from the ground truth of Dhaka, that the boundaries of most informal areas are quite obvious, usually in the forms of segmented patches or clusters, rather than continuous areas. They are mainly distributed near rivers, lakes and other waters, around airports, and at the junction of urban and rural areas. But there are also informal areas in densely populated urban areas where commercial and residential communities live.
There are large formal areas near Ramna Thana and Lalmatia.

\subsection{DeepLabV3+  for semantic segmentation}
\begin{figure}
\includegraphics[scale=0.62]{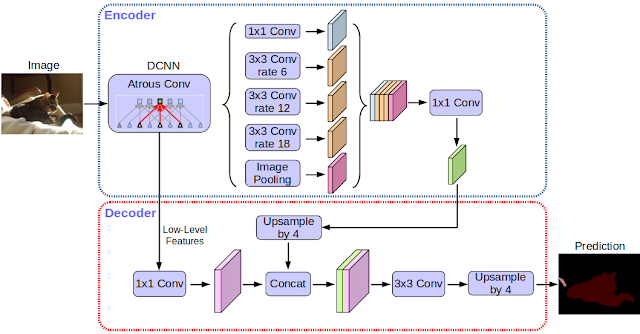}
\caption{Architecture of DeeplabV3+}
\label{fig6}
\end{figure}

DeepLabV3+(Figure \ref{fig6}) is the state of the art model for semantic segmentation. This model has some unique modules that would help efficient and correct segmentation of satellite images. It uses ImageNet’s pre-trained ResNet-101 with atrous convolutions as its main feature extractor. In the modified ResNet model, the last ResNet block uses atrous convolutions with different dilation rates. Atrous convolution with different dilation rate enables the model to extract image features in different scales. Moreover, it uses Atrous Spatial Pyramid Pooling (ASPP)and bilinear upsampling for the decoder module on top of the modified ResNet block. ASPP helps the model to see satellite images from a different point of view. Analyzing the satellite images from a different view is crucial for good segmentation. 
ASPP helps the DeepLabV3+ model to extract features from different levels: ranging from very low-level pixel data to high-level contextual information. 
Therefore, the ASPP is the major contributing module to our semantic segmentation.

The overall architecture has two main modules: encoder and decoder.  The encoder is based on an output stride (ratio of the original image size to the size of the final encoded features) of 16. In the encoder, the ResNet 101 convolution layers are used. Moreover, the ASPP has been exploited in the encoder to fuse features in different scales. In the decoder, instead of using bilinear upsampling with a factor of 16, the encoded features are first upsampled with a factor of 4 and concatenated with corresponding low-level features from the encoder module having the same spatial dimensions. In the decoder, before concatenating, $1{\times}1$ convolutions are applied on the low-level features to reduce the number of channels. After concatenation, a few $3{\times}3$ convolutions are applied and the features are upsampled by a factor of ${4}$. This gives the output of the same size as that of the input image.

\subsection{Creating training data with $70\%$ overlap}
The convolution layers of the DeeplabV3+ are pre-trained with Imagenet. However, the Imagenet is the data-set of objects. In order to use the model for satellite image segmentation, we need to fine-tune the model weights by applying the sufficient training set. 

Therefore, we split the ground truth and the input image into two parts. One part for the training and the other for the testing. The training part consists of the $70\%$ of the Dhaka image that includes the upper part of the Dhaka city. Now before feeding the image into the network, some required pre-processing of the data is needed. 

In the first step, both the input image and its target image needs to be grid into $513{\times}513$ because this is the required width and height to feed into DeepLabV3+. But this grid will be done using $70\%$ overlap of the previous image. This 70\% overlap grid will be done both vertically and horizontally. The reason for overlapping is to increase the training data. Doing the overlap increases the data drastically.
Afterward, mapping of the input image is done using the ground truth image data. Mapping the unrecognized image helps to prevent misclassification, we projected the unrecognized part on the input image. This will now classify the black region as only unrecognized. After all the pre-processing is done, the images are feed into the DeepLabV3+.
\subsection{Training Procedure}
\subsubsection{Training}
When the training data are ready we feed those data into the model. The model that we used in our training is DeeplabV3+. 
We used resnet\cite{He2016} and xception \cite{chollet2017xception} as backbone.
First, the model takes an image as input where the dimension is $513\, {\times}\, 513\, {\times}\, 3$. It goes into the backbone where the batch normalization is done. The image data then downsampled with respect to the given strides and dilations. Then the output of the backbone is fed to the aspp where the atrous convolution happens. We used 4 aspp modules and the output is pooled by global average pooling at the end of the aspp which is later interpolated in bilinear interpolation and sent to the decoder part. From the encoder part, aspp sends the low-level features to the decoder by doing $1\, {\times}\, 1$ convolution and backbone sends the output and these two are concatenated. Later the images are upsampled.

\subsubsection{Hyperparameters}
Then we calculate weights over the total number of classes for balancing the class weights. 
We used Stochastic Gradient Descent (SGD) as our optimizer. We have $5$ classes in our training dataset. We train the model with a learning rate of $0.007$, the number of epochs $26$.

\subsection{Testing Procedure}
For the test data, we use the lower $30\%$ of the original image as shown in Figure \ref{fig4}. The same method of pre-processing is done but in this case, during grid, no overlapping of the image is done since it would be only used for testing.

After the testing is done, DeepLabV3+ generates output images. Those output images are later converted to RGB images for both calculation and visual representation. All the images are then stitched back to their original dimension.


Both the output image and ground truth image are used to generate a confusion matrix. This matrix is used to generate results.

Calculating the confusion matrix gives us the accurate accuracy of our classification model whether it is getting right and the types of errors it is making. We compare this output image with our original ground truth image, in this case, we use the pixel by pixel accuracy which is known as pixel-wise classification.
We use a confusion matrix to describe the performance of the model and derive other performance metrics from it.
\\
\\
Accuracy: Accuracy is calculated by summing the number of correctly classified pixel values and divided by the total number of pixel values.

IoU : IoU is calculated by dividing the area of overlap by the area of union. We compute the area of overlap in the numerator and the area of union in the denominator.

Recall: Recall is a measure of completeness; what percentage of positive pixels are predicted as such.

Precision: Precision is the measure of exactness; what percentage of pixels are predicted as such.

The equation of these measurements given below:
\begin{itemize}
    \item Accuracy: $ \frac{tp + tn}{tp+tn+fp+fn}$\\
    \item IoU score: $\frac{target\; {\cap}\; prediction}{target\; {\cup}\; prediction}$\\
    \item Recall: $\frac{tp}{tp+fn}$\\
    \item Precision: $\frac{tp}{tp+fp}$
\end{itemize}
where tp = true positive, tn = true negative, fp = false positive, and fn = false negative.

\begin{figure*}[!ht]
\centering
\includegraphics[scale=.850]{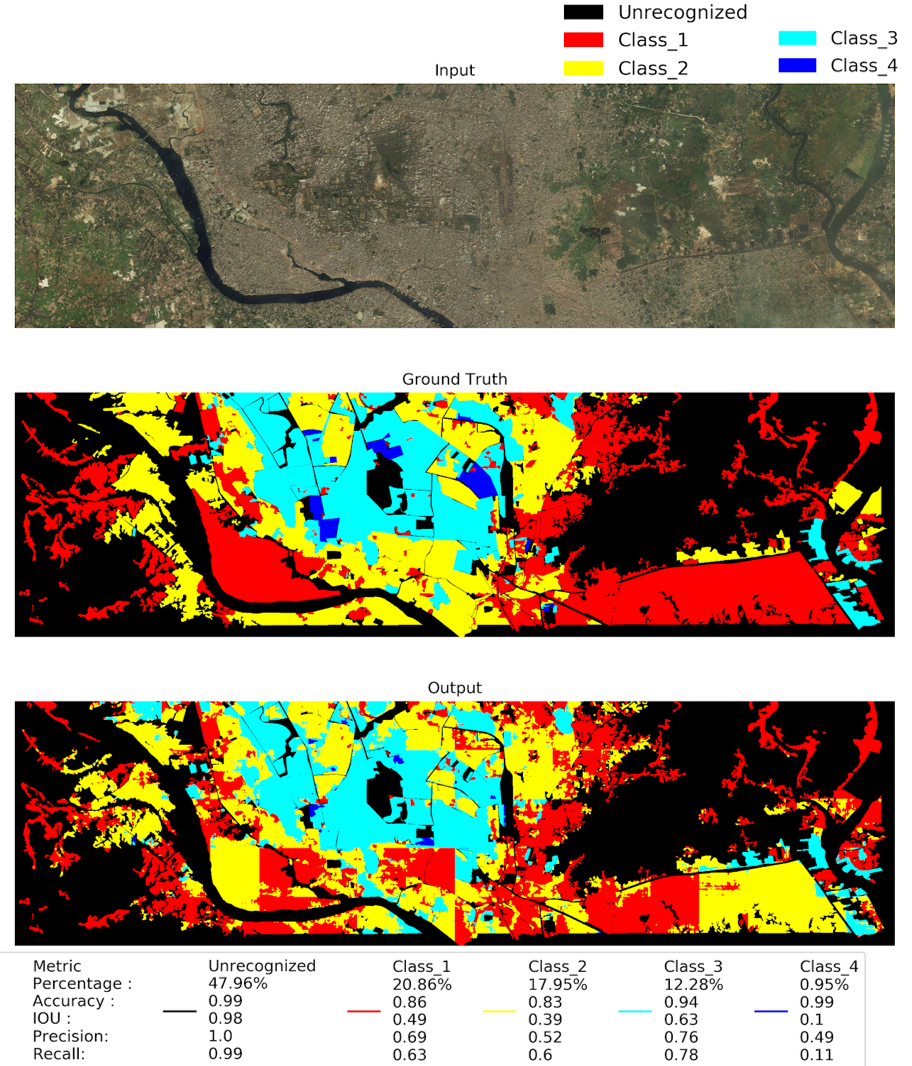}
\caption{Top: Test image, Middle: Ground truth, Bottom: Predicted output - Testing with $30\%$ bottom part of Dhaka}
\label{fig8}
\end{figure*}

\section{Result}

Figure \ref{fig8} shows the performance of the segmentation. The top sub-image is the test image that has been used as the input to the trained model. The bottom image is the output from the segmentation model. The middle one is the ground truth image where different colors represent different building classifications. 

It can be easily noticed that the model can segment the Highly informal area (Class-1) with $86\%$ accuracy. The precision and recall are also very high.  We can also see that the Highly Formal area has also $99\%$ accuracy. Moderately Informal area (Class-2) is a little bit difficult and error-prone compared to other classes.  In contrast, moderately formal areas can be segmented effectively with 94\% accuracy with reasonable precision and recall. Therefore, the deep learning model can categorize urban areas with an average of $90.5\%$. So, the automatic classification by the deep learning model facilitates large scale classification over different cities in developing countries.

\section{Conclusion}
In this study, we have classified the urban environment into $16$ classes using a novel method. The method classifies an urban space using two axes of architectural form and urbanization process and mapped a large range $(50\, km\, {\times}50\, km)$ of urban environment based on satellite images incorporating human visual recognition method. Deep learning method is used to automate the entire classification process. This classification method allows us to match spatial knowledge with urban location information of possible locations, combine topography, area, population density, income, and education data, and adopt a more systematic and consistent method to locate the urban population globally. For semantic segmentation and automatic classification, we have used Google's DeeplabV3+. satellite image encompassing $70\%$ of the urban Dhaka was used to train the model and the remaining $30\%$ was used for testing. We observe that the segmentation model can segment with $75\%$ accuracy and $60\%$ Mean Intersection over Union (mIoU). Our proposed methods should be a valuable contribution to urban geography and urban poverty research and pave the way to conduct a cost-effective computational approach to conduct a morphological catalog of the global urban environment.

\section*{Acknowledgment}
This paper is a result of the project "SmartEGOV: Harnessing EGOV for Smart Governance (Foundations, methods, Tools) / NORTE-01-0145-FEDER-000037", supported by Norte Portugal Regional Operational Programme (NORTE 2020), under the PORTUGAL 2020 Partnership Agreement, through the European Regional Development Fund (EFDR).
It was also partially funded by ICT Division of ICT Ministry, Bangladesh.

\bibliographystyle{unsrt}
\bibliography{journal_ref}

\begin{thebibliography}{10}

\bibitem{UnitedNations2015}
{}.
\newblock United nations,transforming our world: The 2030 agenda for
  sustainable development. technical report.
\newblock {\em {}}, 2015.

\bibitem{Tingzon_I_2019}
I.~Tingzon, A.~Orden, K.~Go, S.~Sy, V.~Sekara, Ingmar Weber, M.~Fatehkia,
  Manuel García-Herranz, and D.~Kim.
\newblock Mapping poverty in the philippines using machine learning, satellite
  imagery, and crowd-sourced geospatial information.
\newblock {\em ISPRS - International Archives of the Photogrammetry, Remote
  Sensing and Spatial Information Sciences}, XLII-4/W19:425--431, 12 2019.

\bibitem{Roy1997}
Prabal~Kanti Roy and Seiji Hayakawa.
\newblock {Characteristics of Air Pollution Condition in Dhaka City}.
\newblock {\em Journal of Agricultural Meteorology}, 1997.

\bibitem{Tauben2016}
Hannes Taubenböck, Ines Standfuß, Martin Klotz, and Michael Wurm.
\newblock The physical density of the city—deconstruction of the delusive
  density measure with evidence from two european megacities.
\newblock {\em ISPRS International Journal of Geo-Information}, 5:206, 11 2016.

\bibitem{Herold2005}
Martin Herold, Helen Couclelis, and Keith Clarke.
\newblock "the role of spatial metrics in the analysis and modeling of urban
  land use change".
\newblock {\em Computers Environment and Urban Systems}, pages 369--399, 07
  2005.

\bibitem{Huang2007}
Jingnan Huang, X.~X. Lu, and Jefferey~M. Sellers.
\newblock {A global comparative analysis of urban form: Applying spatial
  metrics and remote sensing}.
\newblock {\em Landscape and Urban Planning}, 2007.

\bibitem{DBLP:journals/corr/abs-1806-00894}
Barak Oshri, Annie Hu, Peter Adelson, Xiao Chen, Pascaline Dupas, Jeremy
  Weinstein, Marshall Burke, David~B. Lobell, and Stefano Ermon.
\newblock Infrastructure quality assessment in africa using satellite imagery
  and deep learning.
\newblock {\em CoRR}, abs/1806.00894, 2018.

\bibitem{afrobarometer}
Afrobarometer, 2014, round 6 survey manual, (2014).

\bibitem{DBLP:journals/corr/XieJBLE15}
Sang~Michael Xie, Neal Jean, Marshall Burke, David~B. Lobell, and Stefano
  Ermon.
\newblock Transfer learning from deep features for remote sensing and poverty
  mapping.
\newblock {\em CoRR}, abs/1510.00098, 2015.

\bibitem{Piaggesi_2019_CVPR_Workshops}
Simone Piaggesi, Laetitia Gauvin, Michele Tizzoni, Ciro Cattuto, Natalia Adler,
  Stefaan Verhulst, Andrew Young, Rhiannan Price, Leo Ferres, and Andre
  Panisson.
\newblock Predicting city poverty using satellite imagery.
\newblock In {\em The IEEE Conference on Computer Vision and Pattern
  Recognition (CVPR) Workshops}, June 2019.

\bibitem{Rana2011}
Md~Masud~Parves Rana.
\newblock Urbanization and sustainability: Challenges and strategies for
  sustainable urban development in bangladesh.
\newblock {\em Environment Development and Sustainability}, 13:237--256, 02
  2011.

\bibitem{kit2013}
Oleksandr Kit, Matthias Lüdeke, and Diana Reckien.
\newblock Defining the bull's eye: Satellite imagery-assisted slum population
  assessment in hyderabad, india.
\newblock {\em Urban Geography}, 34:413--424, 05 2013.

\bibitem{UN-Habitat2016}
UN-Habitat.
\newblock {\em {Urbanization and development: emerging futures. World cities
  report 2016}}.
\newblock UN-Habitat, 2016.

\bibitem{Siddiqy_2017}
Md~Siddiqy.
\newblock Urban environment and major challenges in sustainable development:
  Experience from dhaka city in bangladesh.
\newblock {\em South East Asia Journal of Public Health}, 7:12, 12 2017.

\bibitem{Barke2018}
Michael Barke.
\newblock {\em The Importance of Urban Form as an Object of Study}, pages
  11--30.
\newblock {}, 01 2018.

\bibitem{Karrholm2008}
Mattias K{\"a}rrholm.
\newblock The territorialization of a pedestrian precinct in malm{\"o}:
  Materialities in the commercialisation of public space.
\newblock {\em Urban Studies}, 45(9):1903--1924, 2008.
\newblock accepted for publication in 2008. The information about affiliations
  in this record was updated in December 2015. The record was previously
  connected to the following departments: Building Functions Analysis
  (011037000).

\bibitem{Lynch1984}
Kevin Lynch.
\newblock {Good city form}, 1984.

\bibitem{samuels2012urban}
Ivor Samuels, Phillippe Panerai, Jean Castex, and Jean~Charles Depaule.
\newblock {\em Urban forms}.
\newblock Routledge, 2012.

\bibitem{Oliveira2018}
Leise Oliveira, Betty Barraza, Bruno Bertoncini, Cassiano Isler, Dannúbia
  Pires, Ellen Madalon, Jessica Lima, Jose Vidal~Vieira, Leonardo Meira, Lilian
  Bracarense, Renata Bandeira, Renata Oliveira, and Suellem Ferreira.
\newblock An overview of problems and solutions for urban freight transport in
  brazilian cities.
\newblock {\em Sustainability}, 10:1233, 04 2018.

\bibitem{Tauben2018}
Hannes Taubenböck, Nicolas Kraff, and Michael Wurm.
\newblock The morphology of the arrival city - a global categorization based on
  literature surveys and remotely sensed data.
\newblock {\em Applied Geography}, 92, 02 2018.

\bibitem{Jain2007}
M.~Shashi and Kamal Jain.
\newblock {Use of Photogrammetry in 3D modeling and visualization of
  buildings}.
\newblock {\em Journal of Engineering and Applied Sciences}, 2007.

\bibitem{gueguen2014classifying}
Lionel Gueguen.
\newblock Classifying compound structures in satellite images: A compressed
  representation for fast queries.
\newblock {\em IEEE Transactions on Geoscience and Remote Sensing},
  53(4):1803--1818, 2014.

\bibitem{Baud2010}
Isa Baud, Monika Kuffer, Karin Pfeffer, Richard Sliuzas, and Sadasivam
  Karuppannan.
\newblock {Understanding heterogeneity in metropolitan india: The added value
  of remote sensing data for analyzing sub-standard residential areas}.
\newblock {\em International Journal of Applied Earth Observation and
  Geoinformation}, 2010.

\bibitem{Bruzzone2006}
Lorenzo Bruzzone and Lorenzo Carlin.
\newblock {A multilevel context-based system for classification of very high
  spatial resolution images}.
\newblock {\em IEEE Transactions on Geoscience and Remote Sensing}, 2006.

\bibitem{Cheng2015}
Gong Cheng, Junwei Han, Kaiming Li, Zhenbao Liu, Shuhui Bu, and Jinchang Ren.
\newblock Effective and efficient midlevel visual elements-oriented land-use
  classification using vhr remote sensing images.
\newblock {\em IEEE Transactions on Geoscience and Remote Sensing}, 53:1--12,
  07 2015.

\bibitem{DBLP:journals/corr/abs-1709-00308}
John~E. Ball, Derek~T. Anderson, and Chee~Seng Chan.
\newblock A comprehensive survey of deep learning in remote sensing: Theories,
  tools and challenges for the community.
\newblock {\em CoRR}, abs/1709.00308, 2017.

\bibitem{Nogueira_2017}
Keiller Nogueira, Otávio~A.B. Penatti, and Jefersson~A. dos Santos.
\newblock Towards better exploiting convolutional neural networks for remote
  sensing scene classification.
\newblock {\em Pattern Recognition}, 61:539–556, Jan 2017.

\bibitem{Abs2019}
Abu Bakar~Siddik Nayem, Anis Sarker, Ovi Paul, Ali Amin, Ashraful Amin, and
  AKM~Mahbubur Rahman.
\newblock {LULC Segmentation of RGB Satellite Image Using FCN-8}.
\newblock {\em {}}, 2019.

\bibitem{Andries2018}
Ana Andries, Stephen Morse, Jim Lynch, Emma Woolliams, J~Fonweban, and R.J.
  Murphy.
\newblock Translation of remote sensing data into sustainable development
  indicators (isdrs24 conference messina italy june 2018).
\newblock In {\em {}}, 06 2018.

\bibitem{10.5555/3298023.3298229}
Jiaxuan You, Xiaocheng Li, Melvin Low, David Lobell, and Stefano Ermon.
\newblock Deep gaussian process for crop yield prediction based on remote
  sensing data.
\newblock In {\em Proceedings of the Thirty-First AAAI Conference on Artificial
  Intelligence}, AAAI’17, page 4559–4565. AAAI Press, 2017.

\bibitem{Mahabir2018}
Ron Mahabir, Andrew Crooks, Anthony Stefanidis, Arie Croitoru, and Peggy
  Agouris.
\newblock A critical review of high and very high-resolution remote sensing
  approaches for detecting and mapping slums: Trends, challenges and emerging
  opportunities.
\newblock {\em Urban Science}, 2, 01 2018.

\bibitem{Benza2016}
Magdalena Benza, John Weeks, Douglas Stow, David Lopez-Carr, and Keith Clarke.
\newblock A pattern-based definition of urban context using remote sensing and
  gis.
\newblock {\em Remote Sensing of Environment}, 183:250--264, 09 2016.

\bibitem{DBLP:journals/corr/abs-1708-09086}
Caleb Robinson, Fred Hohman, and Bistra Dilkina.
\newblock A deep learning approach for population estimation from satellite
  imagery.
\newblock {\em CoRR}, abs/1708.09086, 2017.

\bibitem{UCmerced2010}
Yi~Yang and Shawn Newsam.
\newblock Bag-of-visual-words and spatial extensions for land-use
  classification.
\newblock In {\em {}}, pages 270--279, 01 2010.

\bibitem{cci2013}
Rainer Hollmann, Chris~J Merchant, Roger Saunders, Catherine Downy, Michael
  Buchwitz, Anny Cazenave, Emilio Chuvieco, Pierre Defourny, Gerrit de~Leeuw,
  Ren{\'e} Forsberg, et~al.
\newblock The esa climate change initiative: Satellite data records for
  essential climate variables.
\newblock {\em Bulletin of the American Meteorological Society},
  94(10):1541--1552, 2013.

\bibitem{Jean790}
Neal Jean, Marshall Burke, Michael Xie, W.~Matthew Davis, David~B. Lobell, and
  Stefano Ermon.
\newblock Combining satellite imagery and machine learning to predict poverty.
\newblock {\em Science}, 353(6301):790--794, 2016.

\bibitem{digitalglobe}
Digitalglobe, https://www.digitalglobe.com/.

\bibitem{eod}
Eod (encuesta origen destino de viajes), http://www.sectra.gob.cl/encuestas
  movilidad/ encuestas movilidad.htm.

\bibitem{marshall2009cities}
Stephen Marshall.
\newblock {\em Cities design and evolution}.
\newblock Routledge, Abingdon, Oxon New York, NY, 2009.

\bibitem{He2016}
Kaiming He, Xiangyu Zhang, Shaoqing Ren, and Jian Sun.
\newblock {Deep residual learning for image recognition}.
\newblock In {\em Proceedings of the IEEE Computer Society Conference on
  Computer Vision and Pattern Recognition}, 2016.

\bibitem{chollet2017xception}
Fran{\c{c}}ois Chollet.
\newblock Xception: Deep learning with depthwise separable convolutions.
\newblock In {\em Proceedings of the IEEE conference on computer vision and
  pattern recognition}, pages 1251--1258, 2017.

\end{thebibliography}

\end{document}